\documentclass[sigconf]{acmart}
\AtBeginDocument{%
  }

\setcopyright{none}
\renewcommand\footnotetextcopyrightpermission[1]{}
\acmConference[ICAIF '26]{The 7th ACM International Conference on AI in Finance}{November 14--17, 2026}{Milan, Italy}

\begin{document}

\title{Confidence Estimation for Financial Vision-Language Models in Chart and Document Understanding}

\author{Reza Khanmohammadi}
\orcid{}
\email{khanreza@msu.edu}
\affiliation{%
  \institution{Michigan State University}
  \city{East Lansing}
  \state{Michigan}
  \country{USA}}
\authornote{Corresponding author}

\author{Simerjot Kaur}
\email{simerjot.kaur@jpmchase.com}
\affiliation{%
  \institution{JPMorgan AI Research}
  \city{New York}
  \state{New York}
  \country{USA}}

\author{Charese H. Smiley}
\email{charese.h.smiley@jpmchase.com}
\affiliation{%
  \institution{JPMorgan AI Research}
  \city{New York}
  \state{New York}
  \country{USA}}

\author{Ivan Brugere}
\email{ivan.brugere@jpmchase.com}
\affiliation{%
  \institution{JPMorgan AI Research}
  \city{New York}
  \state{New York}
  \country{USA}}

\author{Mohammad M. Ghassemi}
\email{ghassem3@msu.edu}
\affiliation{%
  \institution{Michigan State University}
  \city{East Lansing}
  \state{Michigan}
  \country{USA}}

\renewcommand{\shortauthors}{Khanmohammadi et al.}

\begin{abstract}
Large vision-language models (LVLMs) are increasingly used to read financial charts, tables, and documents, where a single misread figure can move a decision and the most authoritative-looking answer is sometimes one the model produced without reading the exhibit. The operational question is therefore trust, not accuracy: which answers can be acted on, and which escalated to a reviewer. We evaluate seven confidence estimators, three inference-only and four trained internal probes, across five open-weight LVLMs and four conditions from three financial visual question-answering benchmarks, one bilingual; every probe is trained only on natural images and applied to finance without adaptation, so the results measure out-of-distribution transfer. Three findings hold. First, the scarce property is calibration, not ranking: the inference baselines rank correct above incorrect answers competitively but are badly overconfident, calibration error far above what a threshold can tolerate, and only the trained probes produce a thresholdable score. Second, reliability is structured rather than global, along two axes a practitioner can read directly: the best estimator shifts with both model and task, none leading more than eight of twenty (model, condition) cells, and a controlled bilingual contrast exposes an apparent language robustness as a composition artifact that dissolves once models are read one at a time. Third, cast as deferral under an error budget, how much can be safely automated is set first by the model's competence and only narrowed by its confidence, so deferral clears a real share of the easiest condition and almost none of the hardest, near zero at a strict 5\% budget. Two trained probes carry the calibration a deferral policy needs, and among them only the grounding-aware one lowers its confidence on answers a model gives without using the figure, separating detected non-grounding from a fluent guess.
\end{abstract}


\maketitle

\section{Introduction}
\label{sec:intro}
\noindent\textbf{In financial automation the binding question is when to trust an answer, not how often the model is right.}
Reading and summarizing charts, tables, and documents is a large and repetitive part of financial analysis, and an LVLM can potentially clear many of the routine figure-based questions and route the rest to an expert. What makes that safe is not how often the model is right on average but whether its right answers can be told apart from its wrong ones. Aggregate accuracy alone cannot do this: a model that is 80\% accurate is wrong on twenty of every hundred answers, and if all hundred carry the same confidence, the expert cannot tell which twenty to catch and must re-check every one, leaving no work actually automated. And that is the observed behavior, since language models are systematically overconfident, reporting nearly the same confidence whether right or wrong~\cite{dang2026instinct}. The operational question is therefore trust: which answers can be acted on, and which must be escalated to an expert.

\noindent\textbf{The harder problem in the multimodal setting is an answer the model gives without using the figure.}
An LVLM prepends visual tokens to a language backbone, but architectural access to the exhibit does not make the answer depend on it: attention heads attend far more strongly to text than to visual tokens, and answer distributions are often nearly unchanged when the image is replaced by a blank frame~\cite{Woo2025AVISC,li2024referencefree}. A model can therefore answer a question about a price chart, an earnings table, or a filing from language priors alone, with the figure contributing nothing, a documented weakness of vision-language models~\cite{li2024referencefree}. This is the worst case for a reviewer, because the answer is not faulty reasoning about the figure but fluent text that never consulted it, delivered with the same conviction as a grounded one, so nothing on the surface marks it as the answer to distrust. We study open-weight LVLMs for a related practical reason: financial data is frequently sensitive and processed on private infrastructure, and several of the estimators we examine read model internals that closed APIs do not expose.

\noindent\textbf{Confidence estimation for financial LVLMs has not been studied on its own.}
Confidence estimators carry useful signal in the LVLM setting~\cite{li2024referencefree}, but they are typically evaluated for discrimination under roughly uniform error cost, whereas deployment imposes a different constraint: draw a principled boundary between what can be automated and what must be escalated while holding error on the automated fraction below a tolerated bound~\cite{kompa2021}. No prior work, to our knowledge, studies confidence estimation for financial LVLMs as its own problem, across the range of figure types and task formats that financial analysis actually spans and across more than one language, under the transfer conditions a practitioner faces, where a signal learned on ordinary images must hold on a chart it has never seen.

\noindent\textbf{This study.}
We evaluate seven confidence estimators across five open-weight LVLMs on three financial VQA benchmarks spanning broad financial reasoning (FinMME), chart understanding (FinChart-Bench), and bilingual financial-document analysis (MME-Finance), reported as four conditions. Every trainable probe is trained only on general-domain natural images and applied to finance without adaptation, so the results test transfer rather than in-domain fitting. We frame the comparison as automate-or-escalate under an error budget: a case is automated only when its confidence clears a threshold, and the rest are deferred to a reviewer. The picture is consistent and cautionary. Standard metrics are poor guides to which estimator to trust, neither a single estimator nor a single model is a safe default, and reliability resolves along two axes a practitioner can read directly, the task being performed and the language it is performed in, both tracking how capable the base model is to begin with.

\noindent\textbf{Contributions.}
First, a systematic out-of-distribution study of confidence estimation for financial LVLMs: seven estimators across five open-weight models and four financial conditions under one deployment-oriented, transfer-only protocol. Second, evidence that confidence reliability in finance is structured along two observable axes rather than being a property of a model or an estimator alone: it varies with the task operation, from figure lookup to numerical reasoning to open-ended captioning, and a controlled bilingual contrast on identical tasks separates a harder task from a weaker model, exposing an apparent language robustness as a composition effect. Third, a deployment account in the language of bounded-error automation: how much can be safely delegated is set first by base-model competence, with the confidence layer adding deployable yield over a zero-cost softmax baseline chiefly where the model is weak, so deferral recovers a real share of the easiest condition and almost none of the hardest. The two trained internal probes are the only estimators whose scores a threshold can trust on calibration across conditions, and among them the grounding-aware one adds a property the others lack: on the cases where the model answers without using the figure, it alone lowers its confidence, separating detected non-grounding from a fluent guess.
\vspace{-10pt}
\section{Related Work}
\label{sec:related}
\noindent\textbf{Calibration and discrimination are distinct, and only calibration supports deferral.}
A confidence estimator attaches to each prediction a score meant to express how likely it is correct, and two distinct properties decide whether that score is usable. A score is \emph{calibrated} when its value matches empirical accuracy, so answers assigned 0.8 confidence are correct about 80 percent of the time, and \emph{discriminative} when higher scores are more often correct than lower ones, whatever the absolute scale. A deferral policy needs calibration, because only a calibrated score lets a fixed threshold map to a controlled error rate; discrimination alone merely orders cases. The two can come apart, and this study asks whether either, learned on ordinary images, survives the move to financial figures the probe has never seen~\cite{tempscaling}.

\noindent\textbf{Confidence estimators differ by where they read the signal.}
Prompt-based methods read a self-assessment off the model's own outputs: P(True)~\cite{kadavath}, Self-Probing~\cite{self-probing}, and verbalized-confidence prompting~\cite{tian-etal-2023-just} elicit a verbal or token-level judgment, resting on the model's capacity to introspect on its own computation, which is uneven for multi-step numerical reasoning, while a prompt ensemble instead aggregates confidence across rephrasings of the query, a technique introduced for vision-language-action policies~\cite{promptensemble} that we adapt to VQA. Internal-state probes train a lightweight classifier on hidden activations: SAPLMA~\cite{saplma} reads the post-answer state, and InternalInspector~\cite{internalinspector} learns contrastively across attention, feed-forward, and activation states from all layers. Internal-stability methods read how a representation responds to controlled perturbation (CCPS~\cite{ccps}), and BICR~\cite{bicr} additionally contrasts the representation under the real figure against one under a blanked figure, so its score reflects whether the answer used the image, the only estimator here trained with such a grounding signal; a related line contrasts chain-of-embedding trajectories to expose the same language-prior reliance~\cite{long2026understandinglanguagepriorlvlms}. All were developed for general-domain discrimination under roughly uniform error cost, with a known tension between optimizing calibration and discrimination jointly~\cite{rmcb}. General-domain LVLM confidence has itself been studied, through sampling-based uncertainty~\cite{vl-uncertainty}, functionally-equivalent input sampling~\cite{festa}, and decoupled calibration~\cite{vl-calibration}, but on natural images under roughly uniform error cost; none asks whether such a signal survives the move to financial figures under a bounded-error deferral constraint.

\noindent\textbf{Reliability, not capability, is the weak axis for financial models.}
Strong models remain unreliable on financial tasks: open-book question answering over filings leaves them wrong or refusing on most questions~\cite{financebench}, and a finance-wide trustworthiness audit places truthfulness among the weakest axes~\cite{fintrust}. The problem carries into multimodal financial models~\cite{fintral} and takes a sharp form on charts, where models answer confidently when the evidence is absent from or contradicts the figure~\cite{charthal}, the ungrounded behavior our study is built to catch. Capability gains do not remove the need to know which answers to trust, which is exactly what a confidence signal must supply.

\noindent\textbf{Selective deferral is the deployment frame, and in finance it has so far been text-only.}
Abstaining when uncertain rather than maximizing aggregate accuracy is the established route to safe automation under an error budget, formalized as selective classification with a risk-coverage tradeoff~\cite{elyaniv,selective-classification} and, in high-stakes settings, as abstention with a second opinion~\cite{kompa2021}, including in visual question answering, where a model does better to abstain than to answer wrong~\cite{reliablevqa}. In finance the framing has been applied only to text: selective prediction over financial text QA~\cite{eclipse}, and financial hallucination detection over text and tables through retrieval and claim verification~\cite{finground}. Ours is the multimodal, internal-state counterpart, scoring a frozen LVLM over financial figures at no added inference cost and judging that score by calibration and bounded-error deferral. We evaluate on FinMME~\cite{finmme}, FinChart-Bench~\cite{finchart}, and the bilingual MME-Finance~\cite{mmefinance}, accuracy benchmarks on which we ask the complementary question of when an answer can be trusted.
\vspace{-10pt}
\section{Study Design}
\label{sec:setup}
\noindent\textbf{Three financial benchmarks, two observable axes of variation.}
We evaluate on three financial VQA benchmarks chosen to span distinct figure types and distinct distances from the natural-image data on which the probes are trained, and built so that confidence quality can be read along two axes a practitioner can see: the task being performed and, within one benchmark, the language it is performed in. FinMME~\cite{finmme} (11{,}099 questions) is a broad financial multimodal benchmark over charts and figures with single-choice, multiple-choice, and numerical formats, the largest and most heterogeneous of the three. FinChart-Bench~\cite{finchart} (7{,}019) is a chart-understanding benchmark with true/false, multiple-choice, and open question-answer items about plotted financial data. MME-Finance~\cite{mmefinance} (2{,}274) is a multi-task financial-document benchmark spanning eleven task types (OCR, entity recognition, spatial awareness, financial knowledge, numerical calculation, and others), bilingual in English (1{,}171) and Chinese (1{,}103); we report the two languages as separate conditions, giving three benchmarks and four evaluation conditions in total. Posing the same MME-Finance tasks in two languages turns the bilingual split into a controlled contrast that holds task content fixed and varies only difficulty, which lets us separate a harder task from a weaker model rather than confounding the two.

\noindent\textbf{Out-of-distribution protocol.}
Every trainable estimator is trained and validated only on general-domain GQA~\cite{gqa} (20{,}000 train, 5{,}000 validation) and evaluated on each financial condition without adaptation, so all financial sets are unseen during training. This isolates the question a practitioner faces: does a confidence signal learned on ordinary images still mean what it says on a chart it has never seen? On MME-Finance-EN we use the same 892-sample English subset as~\cite{bicr}, so the new MME-Finance contributions here are the Chinese condition and the English-versus-Chinese contrast, while FinMME and FinChart-Bench are new evaluations in full.

\noindent\textbf{Models.}
We evaluate five open-weight, instruction-tuned LVLMs spanning 8B to 27B parameters and three vision-encoder lineages: Qwen3-VL-8B, LLaVA-NeXT-13B, InternVL3.5-14B, DeepSeek-VL2, and Gemma-3-27B. All run under identical generation conditions (greedy decoding, 64 new tokens, images downscaled to a maximum of 2{,}048 pixels on the long edge). Base VQA accuracy spans a wide, realistic range and, crucially for the deployment analysis, differs systematically across conditions: pooled over the five LVLMs it is 60.8\% on FinChart-Bench, 44.8\% on MME-Finance-EN, 41.8\% on FinMME, and 34.5\% on MME-Finance-ZH, ordering the conditions by difficulty and by distance from the natural-image training distribution. In this regime no single LVLM is reliable on its own, so how much can be safely delegated depends jointly on the model's competence on a condition and on the confidence layer that decides which answers to trust.

\noindent\textbf{Correctness labels.}
Following standard practice in the confidence-estimation literature, we assign every answer a binary correctness label $y \in \{0,1\}$ with a single \texttt{gpt-5-mini} judge that sees the image, question, gold answer, and generated response and decides semantic equivalence, a protocol shown reliable against human annotators~\cite{calibration-tuning,ccps,rmcb}. Applying one judge uniformly across datasets and models is a deliberate control: differences in measured confidence quality then reflect model behavior rather than grading rules that vary across datasets. To validate the judge on the slice where semantic-equivalence grading is most likely to fail, one author adjudicated a 100-item audit stratified toward numerically-exact and OCR answers; agreement with the judge was 94.8\% (Cohen's $\kappa=0.89$), with no degradation on the hard numerical/OCR slice relative to the rest.

\noindent\textbf{Confidence estimators and metrics.}
The seven estimators group into three inference-only methods that treat the LVLM as a black box (P(True), Self-Probing, Prompt Ensemble) and four trained probes that read or perturb internal representations (SAPLMA, CCPS, InternalInspector, and BICR). Each trained probe uses the architecture reported best in its source paper, is trained on GQA over five seeds $\{23,42,137,2024,3407\}$, and is evaluated with seed-averaged confidence. We report calibration with Expected Calibration Error (ECE, 10-bin) and Brier score, and discrimination with AUROC and average precision (AUCPR); the selective-prediction quantities (risk-coverage and safe yield) are defined in Section~\ref{sec:deploy}.
\section{Results Across Models, Tasks, and Languages}
\label{sec:results}
We report calibration (ECE, Brier) and discrimination (AUROC, AUCPR) with task accuracy on each condition. Headline numbers are pooled over the five LVLMs (every (LVLM, question) pair concatenated, the metric computed once on the shared intersection of questions common to all seven methods per model); trained probes use seed-averaged confidence. Table~\ref{tab:headline} is the central result.

\begin{figure*}[t]
  \centering
  \includegraphics[width=0.8\textwidth]{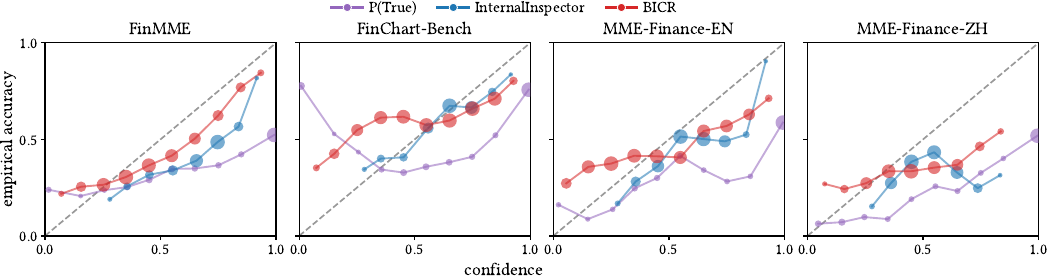}
  \vspace{-10pt}
  \caption{Reliability diagrams pooled across the five LVLMs (dashed line is perfect calibration). Marker size is proportional to the number of samples in each confidence bin. BICR tracks the diagonal closely; P(True) sits above it, issuing high confidence the empirical accuracy does not support, which makes a raw self-report unsafe to threshold for automation.}
  \label{fig:reliability}
  \vspace{-10pt}
\end{figure*}

\begin{table}[t]
  \centering
\caption{Confidence estimation across four financial conditions under out-of-distribution evaluation (probes trained on natural-image GQA, applied to finance without adaptation). Each metric is pooled over the five LVLMs on concatenated (model, question) pairs ($\times100$), so the sample size feeding each metric is about five times the per-model question count reported in the block headers; trained probes use seed-averaged confidence. Calibration: ECE (10-bin), Brier (BS). Discrimination: AUROC, AUCPR. ConfErr is the fraction of a method's errors answered with confidence above 0.8 (lower is better). Best per column in bold.}  \label{tab:headline}
  \vspace{-10pt}
  \resizebox{\linewidth}{!}{%
\begin{tabular}{l cccccc}
\toprule
Method & ECE$\downarrow$ & BS$\downarrow$ & AUROC$\uparrow$ & AUCPR$\uparrow$ & ConfErr$\downarrow$ & ACC$\uparrow$ \\
\midrule
\multicolumn{7}{l}{\textit{FinMME} ($n=11{,}099$ questions/model, base acc 41.8\%)} \\
P(True) & 34.9 & 37.3 & 65.8 & 58.9 & 46.2 & 53.7 \\
Self-Probing & 50.1 & 49.0 & 61.2 & 48.4 & 86.7 & 43.1 \\
Prompt Ens. & 42.1 & 41.2 & 62.1 & 53.0 & 62.6 & 42.2 \\
SAPLMA & 28.4 & 32.4 & 63.5 & 53.8 & 34.2 & 52.3 \\
CCPS & 34.6 & 37.4 & 56.4 & 45.3 & 41.3 & 45.4 \\
InternalInspector & 23.0 & 28.9 & 61.6 & 53.0 & 9.3 & 49.4 \\
\texttt{BICR} & \textbf{8.6} & \textbf{22.4} & \textbf{68.5} & \textbf{63.1} & \textbf{3.7} & \textbf{64.2} \\
\midrule
\multicolumn{7}{l}{\textit{FinChart-Bench} ($n=7{,}019$ questions/model, base acc 60.8\%)} \\
P(True) & 28.5 & 31.9 & 65.0 & \textbf{77.3} & 35.4 & 59.1 \\
Self-Probing & 34.8 & 36.3 & 57.6 & 64.7 & 93.8 & 60.7 \\
Prompt Ens. & 24.8 & 27.4 & \textbf{69.5} & 75.6 & 56.6 & 61.6 \\
SAPLMA & 11.9 & 24.0 & 64.4 & 70.6 & 31.9 & \textbf{65.9} \\
CCPS & 19.5 & 28.0 & 57.8 & 64.5 & 40.8 & 59.6 \\
InternalInspector & \textbf{4.3} & \textbf{22.8} & 62.1 & 69.2 & \textbf{5.9} & 64.5 \\
\texttt{BICR} & 14.9 & 26.2 & 59.0 & 68.8 & 11.6 & 56.2 \\
\midrule
\multicolumn{7}{l}{\textit{MME-Finance-EN} ($n=1{,}171$ questions/model, base acc 44.8\%)} \\
P(True) & 32.5 & 34.2 & \textbf{73.7} & \textbf{71.0} & 47.4 & 58.8 \\
Self-Probing & 49.1 & 48.1 & 62.3 & 52.1 & 91.0 & 45.9 \\
Prompt Ens. & 37.3 & 36.8 & 66.7 & 62.4 & 53.1 & 46.2 \\
SAPLMA & 17.5 & 27.6 & 60.2 & 55.7 & 13.9 & 54.8 \\
CCPS & 30.6 & 36.4 & 53.7 & 47.7 & 43.5 & 48.4 \\
InternalInspector & \textbf{12.4} & 26.0 & 58.9 & 51.2 & \textbf{3.5} & 56.1 \\
\texttt{BICR} & 13.8 & \textbf{25.9} & 61.9 & 56.6 & 7.2 & \textbf{58.9} \\
\midrule
\multicolumn{7}{l}{\textit{MME-Finance-ZH} ($n=1{,}103$ questions/model, base acc 34.5\%)} \\
P(True) & 35.3 & 34.0 & \textbf{77.4} & \textbf{63.7} & 40.4 & \textbf{59.9} \\
Self-Probing & 58.1 & 56.0 & 60.4 & 39.9 & 88.5 & 36.3 \\
Prompt Ens. & 46.0 & 42.7 & 66.3 & 54.6 & 53.3 & 37.0 \\
SAPLMA & 29.7 & 32.2 & 59.3 & 45.5 & 21.6 & 47.9 \\
CCPS & 37.5 & 40.7 & 47.3 & 32.1 & 38.7 & 43.0 \\
InternalInspector & 18.0 & 27.5 & 50.6 & 32.8 & \textbf{0.7} & 50.0 \\
\texttt{BICR} & \textbf{13.7} & \textbf{25.1} & 57.2 & 41.5 & 2.1 & 57.0 \\
\bottomrule
\end{tabular}
  }
  \vspace{-10pt}
\end{table}

\noindent\textbf{What the trained probes buy is calibration, not ranking.}
Across all four conditions the inference-only baselines discriminate competitively but are badly overconfident, posting ECE from 0.25 to 0.58 and placing most of their probability mass far from empirical accuracy (Figure~\ref{fig:reliability}). The failure is starkest for the cheapest self-report: Self-Probing's confident-error rate, the share of its mistakes it still labels with confidence above 0.8 (ConfErr in Table~\ref{tab:headline}), reaches nearly nine in ten, confidence a deferral policy would read as safe. The trained internal probes cut this error by half to an order of magnitude while staying competitive at discrimination, and between them BICR and InternalInspector hold the best calibration on every condition. This is the deployment-relevant property, because a zero-cost logit can rank cases but only a calibrated score can be thresholded into a controlled error rate.

\noindent\textbf{Neither a single estimator nor a single model is a safe default.}
In the pooled view the strongest method already shifts with the domain: BICR owns FinMME on every metric at once (ECE 8.6, BS 22.4, ACC 64.2, AUCPR 63.1, AUROC 68.5); on FinChart-Bench the roles split, with InternalInspector the best calibrated, a Prompt Ensemble the best discriminator (AUROC 69.5), and SAPLMA the most accurate; and on the harder MME-Finance benchmark no internal probe leads ranking at all, P(True) taking AUROC in both languages (73.7 EN, 77.4 ZH). Disaggregated to the twenty (model, condition) cells the lead scatters further: the highest-AUROC method is P(True) in eight cells and BICR in five, with SAPLMA, Self-Probing, and Prompt Ensemble in two each and InternalInspector in one, and no method wins more than eight of twenty. A Friedman test over the seven methods across the cells is significant ($\chi^2=41.1$, $p<10^{-6}$): the methods are separable, but none dominates. A practitioner reading only a pooled leaderboard would pick the wrong tool on most conditions, so confidence estimators for financial LVLMs must be reported per model and per condition rather than collapsed to one number (Table~\ref{tab:significance}).

\begin{table}[t]
  \centering
  \caption{Significance of BICR's discrimination (DeLong test for correlated AUROCs on pooled samples) on the two conditions where it is the top discriminator, and the cross-cell Friedman test over the twenty (LVLM, condition) cells.}
  \label{tab:significance}
  \vspace{-10pt}
  \resizebox{0.9\linewidth}{!}{%
\begin{tabular}{l cc}
\toprule
\texttt{BICR} vs. & $\Delta$AUROC (pts) & DeLong $p$ \\
\midrule
\multicolumn{3}{l}{\textit{FinMME}} \\
P(True) & +2.66 & 0.0e+00$^{***}$ \\
SAPLMA & +4.95 & 0.0e+00$^{***}$ \\
Prompt Ens. & +6.36 & 0.0e+00$^{***}$ \\
InternalInspector & +6.88 & 0.0e+00$^{***}$ \\
Self-Probing & +7.32 & 0.0e+00$^{***}$ \\
CCPS & +12.12 & 0.0e+00$^{***}$ \\
\multicolumn{3}{l}{\textit{MME-Finance-EN}} \\
P(True) & -11.83 & 0.0e+00$^{***}$ \\
Prompt Ens. & -4.83 & 3.0e-06$^{***}$ \\
Self-Probing & -0.45 & 6.7e-01 \\
SAPLMA & +1.69 & 3.9e-02$^{*}$ \\
InternalInspector & +2.96 & 5.4e-05$^{***}$ \\
CCPS & +8.19 & 0.0e+00$^{***}$ \\
\midrule
\multicolumn{3}{l}{Friedman over 20 (LVLM,condition) cells: $\chi^2=41.1$, $p=2.7e-07$} \\
\bottomrule
\end{tabular} 
  }
  \vspace{-10pt}
\end{table}

\begin{figure*}[t]
  \centering
  \includegraphics[width=0.8\textwidth]{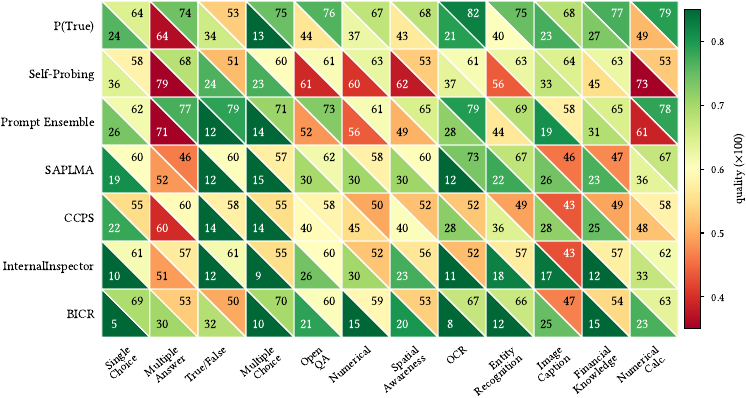}
  \vspace{-10pt}
  \caption{Per financial task-type confidence quality (twelve largest native task and format categories pooled across the benchmarks). Each cell is split diagonally: the upper-right number and color are AUROC (higher and greener is better), the lower-left number is ECE (lower is better). AUROC separability is organized by operation, highest on lookup tasks (OCR, single- and multiple-choice) and lowest on captioning and true/false, whereas calibration tracks the method rather than the task: the four trained probes hold low ECE across nearly every task while the inference-only methods stay miscalibrated throughout. No method is uniformly best across tasks.}
  \label{fig:pertask}
\end{figure*}

\noindent\textbf{Whether a confidence score is trustworthy depends on the task operation, not the benchmark.}
Pooling the twelve largest native task and format categories across the three benchmarks (Figure~\ref{fig:pertask}) shows that separability is organized by the kind of operation the model performs on the figure, while calibration in the same cells tracks the method rather than the task, the trained probes holding low ECE across nearly every operation and the inference-only methods staying miscalibrated throughout. Confidence is most trustworthy on lookup-style tasks, where the best methods reach AUROC near 0.80 on OCR and the high 0.70s on single- and multiple-choice items, and collapses toward chance on open-ended judgment: nearly every method sits at or below 0.69 on image captioning, and most sit near 0.50 on true/false verification, where a fluent guess is hard to tell from a grounded answer. No method owns a row, with P(True) and the Prompt Ensemble taking the most categories while BICR is strongest on single-choice, multiple-choice, and OCR and weakest on captioning and true/false. The operational reading is that a confidence-gated pipeline should follow the task mix, trusting the score first on retrieval and structured-answer operations and last on free-form description.

\noindent\textbf{A controlled bilingual contrast separates a harder task from a weaker model.}
Because MME-Finance poses the same tasks in English and Chinese, the two conditions form a natural experiment that isolates difficulty from competence. The Chinese side is harder, pooled base accuracy 34.5\% against 44.8\% for English, a ten-point gap concentrated in the weaker models. Pooled method rankings look preserved across the two languages, P(True) the top pooled discriminator and BICR the best calibrated on each, but this stability is an artifact of aggregation rather than a property of any estimator: within individual models P(True)'s AUROC falls from English to Chinese in four of the five LVLMs, including ones whose base accuracy barely moves across language, so its steady per-language pooled figures are a composition effect of mixing models with different base rates. P(True) also stays poorly calibrated on Chinese (ECE up to 0.61), so even where it ranks well its score is not thresholdable. The language changes how much can be trusted, not which signal to trust, and the pooled number hides the within-model decline that a deployment on Chinese documents would actually meet.

\noindent\textbf{Where BICR leads, it leads through calibration and a low confident-error rate, not through ranking.}
On FinMME, where BICR is the top discriminator, a DeLong test for correlated AUROCs places it significantly ahead of all six baselines (Table~\ref{tab:significance}); on MME-Finance-EN its edge is partial, ahead of the other trained probes but behind the self-report and the prompt ensemble, the document regime where no internal probe leads ranking. Calibration is the more consistent evidence: the two trained probes hold the best calibration on every condition and rarely fire confidently on a wrong answer, with ConfErr of 2 to 12\% for BICR and as low as 0.7\% for InternalInspector against 35 to 47\% for P(True). A score confidently wrong on up to nearly half its errors cannot be thresholded into a safe policy, which is what the deployment analysis turns on.
\vspace{-10pt}
\section{Deployment: What Can Be Safely Delegated}
\label{sec:deploy}
\begin{figure*}[t]
  \centering
  \includegraphics[width=0.8\textwidth]{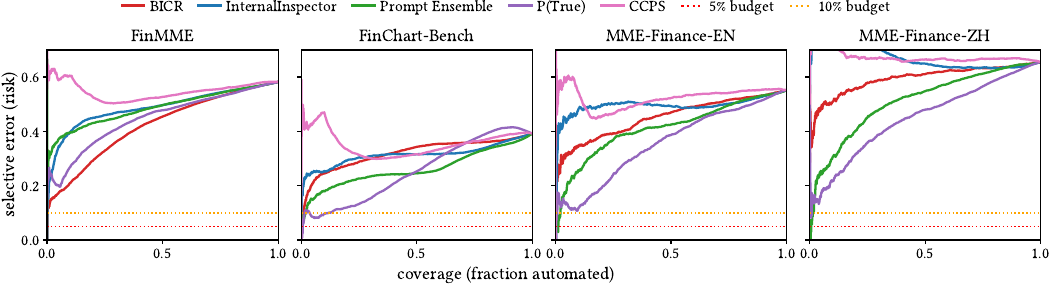}
  \vspace{-10pt}
\caption{Risk-coverage curves pooled across five LVLMs (lower is better). Dotted lines mark the 5\% and 10\% error budgets. FinChart-Bench admits real automation; FinMME admits little; MME-Finance clears the budget only at negligible coverage.}
\label{fig:riskcoverage}
\vspace{-10pt}
\end{figure*}

\noindent\textbf{The deployable question is automate-or-escalate under an error tolerance.}
Aggregate calibration and discrimination do not answer the question an operation faces: how much of this workload can be cleared automatically, and which cases must reach a person? We adopt the standard bounded selective-prediction framing~\cite{selective-classification,elyaniv,kompa2021}, prior apparatus we apply to financial VQA rather than a contribution of this work: automate a case only when its confidence clears a threshold, defer the rest, and judge a method by what it clears under an error budget a deploying desk would set from the cost of a wrong figure. Per (condition, method) we trace the risk-coverage curve, error among automated cases against the fraction automated (Figure~\ref{fig:riskcoverage}, the five most informative methods shown), and read off the \emph{safe yield}, the largest fraction automatable while holding that error at or below a tolerance, which Table~\ref{tab:deploy} reports per model for the three estimators the analysis turns on. Calibration is what makes the threshold meaningful, since a cutoff at 0.8 maps to a controlled error rate only when 0.8 means about 80\% correct.

\begin{figure}[t]
  \centering
  \includegraphics[width=0.75\linewidth]{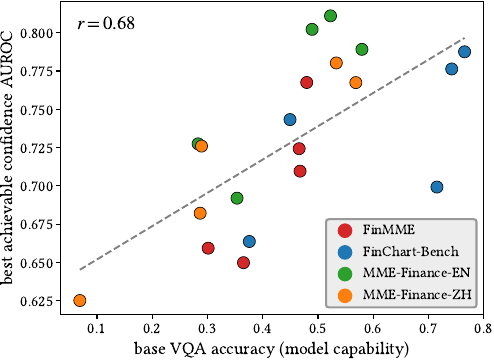}
  \vspace{-10pt}
\caption{Attainable confidence separability rises with base VQA accuracy across the twenty (model, condition) cells ($r=0.68$). Each marker is one cell, colored by condition.}
  \label{fig:capability}
  \vspace{-15pt}
\end{figure}

\noindent\textbf{What can be delegated is set by the model, and the confidence layer earns its yield only where competence runs out.}
Safe yield tracks base-model competence, which tracks each condition's difficulty: across the twenty cells the best attainable confidence AUROC rises with base accuracy ($r=0.68$, Figure~\ref{fig:capability}), so the confidence signal is least reliable exactly where the model is weakest. To ask what the confidence layer adds beyond competence, we compare each method's yield against the model's own answer softmax, a zero-cost ranker that uses no trained layer. The layer's yield advantage is real but concentrated: on the harder conditions it adds a meaningful share (BICR $+8.7$ points on FinMME, P(True) $+17.2$ on MME-Finance-EN over the softmax baseline at the 20\% budget), whereas on the easy FinChart-Bench the free softmax already deploys 39.2\% and no method beats it. Deployable yield under a tuned threshold, then, is set first by the model and improved by a confidence signal chiefly where the model is weak. That comparison rewards ranking under a tuned threshold and is blind to calibration, the fixed-cutoff safety the free softmax lacks and the trained probes supply (Table~\ref{tab:headline}). The confidence layer's contribution is therefore two-part: domain-specific yield where competence is low, and domain-general calibration a zero-cost baseline cannot provide.

\begin{table}[t]
  \centering
\caption{Per-model safe yield under bounded selective prediction for the three deployment-relevant estimators (P(True), InternalInspector, \texttt{BICR}), at the 5\% and 20\% error budgets. Safe yield is the maximum automatable fraction while keeping error among automated cases at or below the budget ($\uparrow$); ``base acc'' is task accuracy. Each block reports the five LVLMs with model-specific operating points, followed by an italic \emph{Pooled} row under a shared threshold. Bold marks the best of the five models per column (excluding \emph{Pooled}). Pooling collapses per-model yield whenever strong and weak models are mixed, with the 1.0\% versus 94.4\% gap on FinChart-Bench the sharpest example.}
\vspace{-5pt}
  \label{tab:deploy}
  \resizebox{\linewidth}{!}{%
\begin{tabular}{l c cc cc cc}
\toprule
 &  & \multicolumn{2}{c}{P(True)} & \multicolumn{2}{c}{InternalInspector} & \multicolumn{2}{c}{\texttt{BICR}} \\
\cmidrule(lr){3-4} \cmidrule(lr){5-6} \cmidrule(lr){7-8}
LVLM & base acc & @5\% & @20\% & @5\% & @20\% & @5\% & @20\% \\
\midrule
\multicolumn{8}{l}{\textit{FinMME}} \\
Qwen3-VL-8B & 48.0 & 0.0 & 1.3 & 0.0 & 19.3 & \textbf{10.5} & \textbf{31.2} \\
LLaVA-NeXT-13B & 30.1 & 0.0 & 0.0 & 0.0 & 0.0 & 0.0 & 2.5 \\
InternVL3.5-14B & 46.6 & \textbf{3.3} & \textbf{16.8} & 0.0 & 0.0 & 2.4 & 12.8 \\
DeepSeek-VL2 & 36.5 & 0.0 & 0.0 & 0.9 & 3.3 & 0.0 & 0.0 \\
Gemma-3-27B & 46.8 & 2.1 & 12.5 & \textbf{2.4} & \textbf{20.4} & 0.0 & 20.5 \\
\cmidrule(l){1-8}
\textit{Pooled} & \textit{41.8} & \textit{0.0} & \textit{5.8} & \textit{0.0} & \textit{0.8} & \textit{0.1} & \textit{9.0} \\
\midrule
\multicolumn{8}{l}{\textit{FinChart-Bench}} \\
Qwen3-VL-8B & 76.5 & 1.9 & 72.0 & \textbf{49.6} & \textbf{94.4} & \textbf{20.9} & 58.1 \\
LLaVA-NeXT-13B & 37.6 & 0.0 & 0.0 & 0.0 & 0.0 & 0.8 & 4.5 \\
InternVL3.5-14B & 74.2 & \textbf{35.5} & \textbf{72.7} & 0.0 & 18.1 & 0.5 & 16.7 \\
DeepSeek-VL2 & 44.9 & 0.0 & 1.6 & 0.0 & 1.1 & 0.0 & 0.0 \\
Gemma-3-27B & 71.5 & 14.3 & 59.8 & 3.8 & 65.5 & 12.1 & \textbf{69.5} \\
\cmidrule(l){1-8}
\textit{Pooled} & \textit{60.8} & \textit{0.6} & \textit{38.7} & \textit{0.0} & \textit{1.0} & \textit{0.3} & \textit{4.4} \\
\midrule
\multicolumn{8}{l}{\textit{MME-Finance-EN}} \\
Qwen3-VL-8B & 57.9 & 0.0 & \textbf{52.8} & \textbf{4.0} & \textbf{27.7} & 0.0 & \textbf{20.4} \\
LLaVA-NeXT-13B & 28.3 & 0.0 & 2.6 & 0.0 & 0.0 & 0.0 & 0.0 \\
InternVL3.5-14B & 52.3 & \textbf{10.3} & 42.3 & 0.0 & 0.0 & 0.0 & 0.0 \\
DeepSeek-VL2 & 35.4 & 0.0 & 5.5 & 0.0 & 0.0 & 0.0 & 0.0 \\
Gemma-3-27B & 48.9 & 0.0 & 37.7 & \textbf{4.0} & 15.1 & 0.0 & 8.5 \\
\cmidrule(l){1-8}
\textit{Pooled} & \textit{44.8} & \textit{0.0} & \textit{21.1} & \textit{0.0} & \textit{0.6} & \textit{0.0} & \textit{0.0} \\
\midrule
\multicolumn{8}{l}{\textit{MME-Finance-ZH}} \\
Qwen3-VL-8B & 56.8 & 0.0 & \textbf{43.7} & 0.0 & \textbf{9.0} & 0.0 & \textbf{14.5} \\
LLaVA-NeXT-13B & 6.9 & 0.0 & 0.0 & 0.0 & 0.0 & 0.0 & 0.0 \\
InternVL3.5-14B & 53.3 & 0.0 & 34.9 & 0.0 & 3.6 & 0.0 & 0.0 \\
DeepSeek-VL2 & 28.7 & 0.0 & 0.0 & 0.0 & 0.0 & 0.0 & 0.0 \\
Gemma-3-27B & 28.9 & 0.0 & 0.0 & 0.0 & 0.0 & 0.0 & 0.0 \\
\cmidrule(l){1-8}
\textit{Pooled} & \textit{34.5} & \textit{0.0} & \textit{8.7} & \textit{0.0} & \textit{0.0} & \textit{0.0} & \textit{0.0} \\
\bottomrule
\end{tabular} 
  }
\end{table}

\noindent\textbf{The pooled verdict is an artifact of mixing models; per model, automation is real but narrow.}
Pooled over the five LVLMs, safe yield at a strict 5\% budget is near zero on every condition (Table~\ref{tab:deploy}), and useful automation appears only at a looser budget on the easier conditions. But the pooled figure understates what a single deployed model can do, because pooling strong and weak models under one threshold erases the strong model's reachable region: on FinChart-Bench at the 20\% budget InternalInspector pools to 1.0\% while clearing 94.4\% on Qwen3-VL alone, and BICR pools to 0.0\% on MME-Finance-EN while clearing 20.4\% on Qwen3-VL. Automation is therefore real but narrow, confined to the strongest models on the easier conditions and vanishing on the weak models and the document tasks, where no operating point carves out a low-error subset once the model is wrong on more than half of cases. The five financial LVLMs we study are not ready to answer chart and document questions on their own at a deployment-grade tolerance; the workable role is confidence-gated automation with an expert signing off every escalation, and within that role the choice of estimator still matters.

\noindent\textbf{A thresholded BICR score is safe to act on because it backs off when the figure is ignored.}
What makes a score actionable is that a threshold maps to a controlled error rate, which BICR most nearly achieves, and its grounding-aware objective transfers to finance. Replacing each chart with a deterministic random image and isolating the figure-invariant subpopulation, the samples whose first generated token does not change under the swap and on which the model demonstrably ignores the figure, BICR is the best- or tied-best-calibrated of the seven estimators on FinMME, MME-Finance-EN, and MME-Finance-ZH, with significantly lower per-sample Brier than every baseline on the first two and than five of six on the third (paired bootstrap, Table~\ref{tab:grounding}). Because those subpopulations are majority incorrect (42\%, 45\%, 27\% correct), the low, well-calibrated confidence reflects detected non-grounding rather than blanket under-confidence. The effect is absent on FinChart-Bench, whose figure-invariant items are majority correct (61\%) and largely answerable from text, so first-token invariance does not isolate grounding there; we report that null as a scope limit of the diagnostic rather than a failure of the method.

\begin{table}[t]
  \centering
\caption{Calibration on the figure-invariant subpopulation (first generated token unchanged when the chart is replaced with a deterministic random image). On this population the model answers without using the figure, and BICR is best- or tied-best-calibrated on genuinely figure-dependent cases. Significance is a paired bootstrap of per-sample Brier against BICR ($\ast$ = 95\% CI excludes zero). Pooled over five LVLMs.}
  \label{tab:grounding}
  \resizebox{\linewidth}{!}{%
\begin{tabular}{l ccccc}
\toprule
Method & mean conf & AUROC & ECE$\downarrow$ & Brier$\downarrow$ & $\Delta$BS vs BICR \\
\midrule
\multicolumn{6}{l}{\textit{FinMME}: invariant $n=14736$, 42\% correct; verdict: SUPPORTED} \\
P(True) & 0.721 & 0.652 & 0.307 & 0.345 & +0.111$^{*}$ \\
Self-Probing & 0.906 & 0.622 & 0.487 & 0.474 & +0.239$^{*}$ \\
Prompt Ens. & 0.821 & 0.640 & 0.397 & 0.390 & +0.155$^{*}$ \\
SAPLMA & 0.743 & 0.618 & 0.320 & 0.350 & +0.115$^{*}$ \\
CCPS & 0.755 & 0.568 & 0.346 & 0.373 & +0.139$^{*}$ \\
InternalInspector & 0.670 & 0.600 & 0.246 & 0.302 & +0.067$^{*}$ \\
\texttt{BICR} & 0.492 & 0.665 & 0.095 & 0.235 & ref \\
\midrule
\multicolumn{6}{l}{\textit{MME-Finance-EN}: invariant $n=2298$, 45\% correct; verdict: SUPPORTED} \\
P(True) & 0.703 & 0.743 & 0.265 & 0.304 & +0.039$^{*}$ \\
Self-Probing & 0.926 & 0.666 & 0.479 & 0.464 & +0.199$^{*}$ \\
Prompt Ens. & 0.799 & 0.660 & 0.350 & 0.353 & +0.089$^{*}$ \\
SAPLMA & 0.591 & 0.553 & 0.200 & 0.299 & +0.034$^{*}$ \\
CCPS & 0.767 & 0.526 & 0.357 & 0.395 & +0.131$^{*}$ \\
InternalInspector & 0.611 & 0.528 & 0.163 & 0.282 & +0.017$^{*}$ \\
\texttt{BICR} & 0.529 & 0.627 & 0.164 & 0.265 & ref \\
\midrule
\multicolumn{6}{l}{\textit{MME-Finance-ZH}: invariant $n=2215$, 27\% correct; verdict: SUPPORTED} \\
P(True) & 0.550 & 0.801 & 0.275 & 0.263 & +0.004 \\
Self-Probing & 0.904 & 0.616 & 0.629 & 0.596 & +0.337$^{*}$ \\
Prompt Ens. & 0.788 & 0.583 & 0.514 & 0.470 & +0.211$^{*}$ \\
SAPLMA & 0.625 & 0.574 & 0.386 & 0.354 & +0.095$^{*}$ \\
CCPS & 0.722 & 0.331 & 0.540 & 0.539 & +0.280$^{*}$ \\
InternalInspector & 0.592 & 0.385 & 0.317 & 0.331 & +0.072$^{*}$ \\
\texttt{BICR} & 0.476 & 0.546 & 0.211 & 0.259 & ref \\
\midrule
\multicolumn{6}{l}{\textit{FinChart-Bench}: invariant $n=8542$, 61\% correct; verdict: null} \\
P(True) & 0.708 & 0.719 & 0.203 & 0.260 & +0.004 \\
Self-Probing & 0.925 & 0.559 & 0.331 & 0.350 & +0.093$^{*}$ \\
Prompt Ens. & 0.864 & 0.708 & 0.257 & 0.282 & +0.026$^{*}$ \\
SAPLMA & 0.775 & 0.670 & 0.177 & 0.255 & -0.001 \\
CCPS & 0.811 & 0.555 & 0.234 & 0.297 & +0.040$^{*}$ \\
InternalInspector & 0.662 & 0.537 & 0.070 & 0.244 & -0.013 \\
\texttt{BICR} & 0.654 & 0.594 & 0.144 & 0.257 & ref \\
\bottomrule
\end{tabular} 
  }
  \vspace{-10pt}
\end{table}
\vspace{-5pt}
\section{Discussion}
\label{sec:discussion}
\noindent\textbf{A deployment threshold consumes calibration, not ranking.}
A score is actionable only if it can be thresholded: when 0.8 means roughly 80\% correct, a cutoff maps to a controlled error rate~\cite{kompa2021}. Discrimination is necessary but not sufficient, because a perfectly ranked yet miscalibrated score yields no safe operating point, which is why the inference-only baselines, competitive at AUROC yet overconfident at ECE 0.25 to 0.58, are non-functional for deferral on the harder conditions. The portable lesson across four financial conditions is that the trained internal probes earn their place on calibration: BICR and InternalInspector are the only methods whose scores a threshold can trust, and BICR's margin is widest on the broad benchmark where the figure is most often necessary.

\noindent\textbf{No single estimator and no single model is a safe default.}
The preferred method shifts with both domain and model: BICR owns FinMME on every metric, a prompt ensemble is the best discriminator on chart understanding, and a raw self-report is both the strongest discriminator and the most automatable estimator on MME-Finance. A practitioner reading only a pooled leaderboard would pick the wrong tool on most conditions, so methods should be reported and selected per model and per task family; the two trained probes are the safe general choice for a thresholded score, and the grounding-aware one is preferable where visual ungroundedness is the failure to guard against.

\noindent\textbf{The honest near-term role is a confidence-gated reviewer, not an autonomous one.}
Selective deferral, not raw accuracy, is what permits safe automation: the near-term opportunity is clearing routine figure questions and prioritizing the rest with a person on every escalation, and the risk is mistaking a score that is merely discriminative, or merely low, for one that is safe. As financial LVLMs improve, the safe-yield frontier will rise unevenly, fastest on the models and domains closest to general training data and slowest on the hardest document tasks, and a well-calibrated estimator, grounding-aware where ungroundedness is the risk, is the signal whose meaning a user can rely on at any point along that frontier.
\section{Limitations}
\label{sec:limitations}
Our probes are trained on a single natural-image source (GQA) and transferred zero-shot, which is the transfer setting we set out to measure rather than the best achievable result; in-domain financial training is left to future work and would test a different question. Correctness labels come from a model judge, standard in this literature but imperfect for numerically exact answers. The per-task analysis pools categories across benchmarks for statistical power, and the deployment numbers use a single shared threshold in the pooled view and a model-specific one per cell, whereas a real system would set the threshold on a held-out slice of the target distribution. The behavioral grounding test uses first-token invariance as a proxy for full-answer invariance, which isolates figure-grounding only where the task is not answerable from text, as the FinChart-Bench null shows.
\section{Conclusion}
\label{sec:conclusion}
We evaluated seven confidence estimators across five open-weight LVLMs and four financial conditions under one out-of-distribution protocol, asking whether a confidence signal learned on ordinary images can tell, on financial figures it has never seen, when an answer should be trusted. Three findings hold. First, the trained probes earn their place on calibration, not ranking: the inference baselines discriminate competitively but are too overconfident to threshold, while only BICR and InternalInspector map to a controlled error rate. Second, reliability is structured, not global: it shifts with model and task, no method winning more than eight of twenty cells, and a bilingual contrast exposes an apparent language robustness as a composition effect, so methods must be selected per model and per task family. Third, how much can be safely delegated is set first by base-model competence ($r=0.68$); against the model's own softmax the confidence layer improves yield chiefly where the model is weak, its domain-general contribution being calibration, not yield. Across all three, only the grounding-aware probe lowers its confidence on answers the model gives without using the figure, separating detected non-grounding from a fluent guess. The five LVLMs we study cannot read charts and documents on their own at a deployment-grade tolerance; the near-term role is confidence-gated automation with an expert on every escalation.

\section*{Acknowledgments}
This work was supported by the JPMorgan Chase AI Research Faculty Research Award. The authors are solely responsible for the contents of this paper; the opinions expressed do not necessarily reflect those of the funding organizations. The authors also acknowledge the use of Large Language Models to assist in polishing the language and grammar of this manuscript.

\section*{Disclaimer}
This paper was prepared for informational purposes by the Artificial Intelligence Research group of JPMorgan Chase \& Co and its affiliates (``JP Morgan''), and is not a product of the Research Department of JP Morgan. JP Morgan makes no representation and warranty whatsoever and disclaims all liability, for the completeness, accuracy or reliability of the information contained herein. This document is not intended as investment research or investment advice, or a recommendation, offer or solicitation for the purchase or sale of any security, financial instrument, financial product or service, or to be used in any way for evaluating the merits of participating in any transaction, and shall not constitute a solicitation under any jurisdiction or to any person, if such solicitation under such jurisdiction or to such person would be unlawful.

\bibliographystyle{ACM-Reference-Format}
\bibliography{references}
\end{document}